 \documentclass[tablecaption=bottom,wcp]{jmlr} 

\usepackage{booktabs}
\usepackage[load-configurations=version-1]{siunitx} 

\usepackage{mathtools}
\usepackage{bm}
\usepackage{caption}
\usepackage{tikz}
\usepackage{tikzscale}
\usetikzlibrary{shapes.geometric,arrows,chains,matrix,positioning,scopes,calc, bayesnet}
\tikzstyle{mybox} = [draw=white, rectangle]

\usepackage{cleveref}

\usepackage{xcolor}

\usepackage{MarkMathCmds}
\newcommand{\obs}{\mathbf{Y}}

\newcommand{\defas}{\triangleq}

\newcommand{\X}{\mathbf{X}}
\newcommand{\Y}{\mathbf{Y}}
\newcommand{\xo}{\mathbf{x}_1}

\newcommand{\xt}{\mathbf{x}_t}
\newcommand{\yt}{\mathbf{y}_t}
\newcommand{\at}{\mathbf{A}_t}
\newcommand{\bt}{\mathbf{b}_t}

\newcommand{\st}{\mathbf{S}_t}
\newcommand{\rft}{\mathbf{f}_t}
\newcommand{\ft}{f(\xt)}

\newcommand{\xtp}{\mathbf{x}_{t+1}}
\newcommand{\bu}{\mathbf{u}}
\newcommand{\muu}{\bm{\mu}_{\mathbf{u}}}
\newcommand{\su}{\mathbf{\Sigma}_{\mathbf{u}}}
\newcommand{\Z}{\mathbf{Z}}
\newcommand{\smu}{\setminus \mathbf{u}}

\newcommand{\calcd}[1][]{\mathrm{d}#1}
\newcommand{\Exp}[2]{\mathbb{E}_{#1}\left[#2\right]}
\DeclarePairedDelimiterX{\dbar}[2]{[}{]}{\,#1\,\delimsize\|\,#2\,}
\newcommand{\KL}[2]{\mathrm{KL} \dbar*{#1}{#2}}

\newcommand{\cft}{\mathbf{C}_f(\xt)}

\newcommand{\cfut}{\mathbf{C}_{f|\vu}(\xt)}

 \usepackage{paralist}
 \usepackage{wrapfig}
 \usepackage{appendix}

\jmlrproceedings{AABI 2018}{1st Symposium on Advances in Approximate Bayesian Inference, 2018}

\title{Non-Factorised Variational Inference in Dynamical Systems}

  \author{\Name{Alessandro Davide Ialongo} \Email{adi24@cam.ac.uk}\\
  \addr University of Cambridge \and Max Planck Institute for Intelligent Systems
  \AND
  \Name{Mark {van der Wilk}} \Email{mark@prowler.io}\\
  \addr PROWLER.io
  \AND
  \Name{James Hensman} \Email{james@prowler.io}\\
  \addr PROWLER.io
  \AND
  \Name{Carl Edward Rasmussen} \Email{cer54@cam.ac.uk}\\
  \addr University of Cambridge
 }

\begin{document}

\maketitle

\vspace{-1.0cm}
\begin{keywords}
Gaussian processes, variational inference, state-space models, time-series, GPSSM.
\end{keywords}

\section{Introduction}

Many real world systems exhibit complex non-linear dynamics, and a wide variety of approaches have been proposed to model them. Characterising these dynamics is essential to the analysis of a system's behaviour, as well as to prediction. Autoregressive models map past observations to future ones and are very common in diverse fields such as economics and model predictive control. State-space models (SSMs), lift the dynamics from the observations to a set of auxiliary, unobserved variables (the latent states \(\{\vx_t\}_{t=1}^T\)) which fully describe the state of the system at each time-step, and thus take the system to evolve as a Markov chain. The ``transition function'' maps a latent state to the next. 
We place a Gaussian process (GP) prior on the transition function, obtaining the Gaussian process state-space model (GPSSM). This Bayesian non-parametric approach allows us to:
\begin{inparaenum}[1)]
\item obtain uncertainty estimates and predictions from the posterior over the transition function,
\item handle increasing amounts of data without the model saturating, and
\item maintain high uncertainty estimates in regions with little data.
\end{inparaenum}


While many approximate inference schemes have been proposed \citep{frigola2013gpssm-pmcmc}, we focus on variational ``inducing point'' approximations \citep{titsias2009}, as they offer a particularly elegant framework for approximating GP models without losing key properties of the non-parametric model. Since a non-parametric Gaussian process is used as the approximate posterior \citep{matthews2016kl}, the properties of the original model are maintained. Increasing the number of inducing points, we add capacity to the approximation, and the quality of the approximation is measured by the marginal likelihood lower bound (or evidence lower bound -- ELBO).

The accuracy of variational methods is fundamentally limited by the class of approximate posteriors, with independence assumptions being particularly harmful for time-series models \citep{turner2010}. Several variational inference schemes that factorise the states and transition function in the approximate posterior have been proposed \citep{frigola2014vgpssm,mchutchon2014nonlinear,ialongo2017gpssm,eleftheriadis2017gpssm}. Here, we investigate the design choices which are available in specifying a non-factorised posterior.

\section{Gaussian Process State Space Models}
Conceptually, a GPSSM is identical to other SSMs. We model discrete-time sequences of observations \(\obs = \left\{\vy_t\right\}_{t=1}^T\), where \(\vy_t \in \Reals^E\), by a corresponding latent Markov chain of states \(\X = \left\{\vx_t\right\}_{t=1}^T\) where \(\vx_t \in \Reals^D\). All state-to-state transitions are governed by the same transition function \(f\). For simplicity, we take the transition \(p(\xtp \given f, \xt)\) and observation \(p(\vy_t \given \vx_t)\) densities to be Gaussians, although any closed form density function could be chosen. Without loss of generality (subject to a suitable augmentation of the state-space) \citep{frigola2015bayesian}, we also assume a linear mapping between \(\vx_t\) and the mean of \(\vy_t \given \xt\) to alleviate non-identifiabilities between transitions and emissions. The generative model is specified by the following equations:
\vspace{-0.1cm}
\begin{equation}
\begin{aligned}
    f &\sim \GP(0, k(\cdot, \cdot)) \, \\
    \vx_{t+1}\given f, \vx_t &\sim \NormDist{f(\vx_t),  \mathbf{Q}} \, \\
    \end{aligned}
\qquad\qquad
\begin{aligned}[c]
    \vx_1 &\sim \NormDist{\mathbf{0}, \mathbf{I}} \, \\
    \vy_t \given \vx_t &\sim \NormDist{\mathbf{C}\vx_t + \mathbf{d}, \mathbf{R}} \, \\
\end{aligned}
\end{equation}
the function values \(\ft\) are given by the GP as:
\vspace{-0.1cm}
\begin{equation}
f(\vx_t) \given \vx_{1:t}, f(\vx_{1:t-1}) \sim \NormDist{\bm{\mu}_{f_t}, \mathbf{C}_{f_t}} \qquad \qquad\quad
f(\vx_{1:t-1}) \defas \left[f(\vx_{1})^\top,\dots, f(\vx_{t-1})^\top\right]^\top
\end{equation}
\vspace{-0.5cm}
\begin{equation*}
\bm{\mu}_{f_t} \defas K_{\xt,\vx_{1:t-1}}K_{\vx_{1:t-1}, \vx_{1:t-1}}^{-1} f(\vx_{1:t-1}) \qquad 
\mathbf{C}_{f_t} \defas K_{\xt,\xt} - K_{\xt,\vx_{1:t-1}}K_{\vx_{1:t-1}, \vx_{1:t-1}}^{-1} K_{\vx_{1:t-1},\xt}
\end{equation*}

\section{Design choices in variational inference}
We want a variational approximation to \(p(\X, f\given \Y)\). We begin by writing our approximate posterior as \(q(\X, f) = q(\X\given f)q(f)\) For \(q(f)\) we choose an inducing point posterior according to \citet{titsias2009} and \citet{hensman2013} and write \(q(f)=p(f^{\smu}\given \vu)q(\vu)\) by splitting up the function into the inducing outputs $\vu$ and all other points $f^{\smu}$. For our \(q(\X\given f)\), we will consider Markovian distributions, following the structure of the exact posterior\footnote{More precisely, the exact conditional posterior is: \(p(\X \given f, \Y) = p(\vx_1 \given f, \Y) \prod_{t=1}^{T-1} p(\xtp\given f(\xt), \Y_{t+1:T})\).}:
\vspace{-0.1cm}
\begin{equation}
    q(\X\given f) = q(\vx_1)\prod_{t=1}^{T-1}q(\vx_{t+1}\given f, \vx_t)
\end{equation}
This allows us to write down the general form of the variational lower bound:
\vspace{-0.1cm}
\begin{align}
    &\log p(\Y) \geq \int q(f, \X) \log \Bigg[\frac{p(f^{\smu}\given \vu)p(\vu)p(\vx_1)\prod_{t=1}^{T-1} p(\vx_{t+1}\given f, \vx_t)}{p(f^{\smu}\given \vu)q(\vu)q(\vx_1)\prod_{t=1}^{T-1} q(\vx_{t+1}\given f, \vx_t)} \prod_{t=1}^T p(\vy_t\given \vx_t)\Bigg] \calcd f \calcd \X \\
    &\phantom{\log p(\Y) \,} = \sum_{t=1}^T \Exp{q(\xt)}{\log p(\vy_t\given \xt)} - \KL{q(\vu)}{p(\vu)} - \KL{q(\vx_1)}{p(\vx_1)} \nonumber \\[-10pt]
    & \phantom{\log p(\Y) \geq} \qquad \qquad - \sum_{t=1}^{T-1} \Exp{q(f, \xt)}{\KL{q(\xtp \given f, \xt)}{p(\xtp \given f, \xt))}} \label{eq:bound}
\end{align}

We are now left to specify the form of \(q(\xo)\), \(q(\vu)\), and \(q(\xtp \given f, \xt)\). For the first two, we choose Gaussian distributions and optimise variationally their means and covariances, while, for the last, several choices are available. We follow the form of the exact filtering distribution (for Gaussian emissions) but treat \(\at,\bt,\st\) as free variational parameters to be optimised (thus approximating the smoothing distribution):
\begin{equation}
    q(\vx_{t+1}\given f, \vx_t) = \NormDist{\vx_{t+1}\given \at \rft + \bt, \st^*}
\end{equation}

\pagebreak
\begin{table}[htpb]
\captionsetup{justification=raggedright, singlelinecheck=false, format=hang}
\centering
\begin{tabular}{lccccc}
                             & \(q(\X \given  f)\)  & $\rft$                         & $\st^*$                   &  sampling \\ \hline
1) Factorised - linear          & \(q(\X)\)                   & $\vx_t$                        & $\st$                    &  $\mathcal{O}(T)$ \\
2) Factorised - non-linear      & \(q(\X)\)                   & $K_{\xt,\Z}K_{\Z,\Z}^{-1}\muu$ & $\st + \at\cft\at^\top$ &  $\mathcal{O}(T)$ \\
3) U-Factorised - non-linear    & \(q(\X\given \vu)\)                   & $K_{\xt,\Z}K_{\Z,\Z}^{-1}\vu$  & $\st + \at\cfut\at^\top$                    &  $\mathcal{O}(T)$ \\
4) Non-Factorised - non-linear  & \(q(\X\given f)\)                   & $f(\vx_t)$                     & $\st$                    &  $\mathcal{O}(T^3)$
\end{tabular}
\caption{Variations of approximate posteriors. \(\at, \bt, \st\) are free parameters in all cases. \(\cft\) is the sparse GP's marginal posterior variance:
\vspace{-0.25cm}
\[
\cft \defas K_{\xt,\xt} + K_{\xt,\Z}K_{\Z,\Z}^{-1} \left(\su - K_{\Z,\Z}\right) K_{\Z,\Z}^{-1} K_{\Z,\xt} 
\]
whereas \(\cfut\) is the conditional variance of \(\ft\given\vu\):
\vspace{-0.25cm}
\[
\cfut \defas K_{\xt,\xt} - K_{\xt,\Z}K_{\Z,\Z}^{-1} K_{\Z,\xt} 
\]}
\end{table}
\vspace{-1.cm}
Each posterior is identified by whether it factorises the distribution between states and transition function, and by whether it is linear in the latent states (i.e. only the first one, which corresponds to a joint Gaussian over the states). The cubic sampling cost associated with the last choice of posterior (``Non-Factorised - non-linear'', i.e. the full GP) derives from having to condition, at every time-step, on a sub-sequence of length \(t \in \{0,\dots, T-1\}\), giving \(T\) operations, each of \(\mathcal{O}(t^2)\) complexity (i.e. updating and solving a \(t \times t\) triangular linear system represented by a Cholesky decomposition). The cost of sampling is crucial, because evaluating and optimising our variational bound requires obtaining samples of \(q(\xt)\) and \(q(f, \xt)\) to compute expectations by Monte Carlo integration. 
We now review two options to side-step the cubic cost associated with the fully non-parametric GP.

\subsection{Dependence on the entire process \(f\) - ``Chunking''}
If we wish to retain the full GP as well as a non-factorised posterior, but the data does not come in short independent sequences, one approach is to ``cut'' the posterior \(q(\X|f)\) into sub-sequences of lengths \(\tau_1, \dots, \tau_n\):
\vspace{-0.2cm}
\begin{align}
    q(\X\given f) = \left[q(\vx_1)\prod_{t=1}^{\tau_1 - 1}q(\vx_{t+1}\given f, \xt)\right]\left[q(\vx_{\tau_1+1})\prod_{t=\tau_1+1}^{\tau_1 + \tau_2 - 1}q(\vx_{t+1}\given f, \xt) \right]\dots \Bigg[\dots\Bigg] \,
\end{align}
for \(\tau_1 = \dots = \tau_n = \tau\) this reduces the cost of sampling to $\mathcal{O}\left(\frac{T}{\tau}\tau^3\right)$. Conditioning (which has cubic cost in the size of the conditioning set) now only needs to extend as far back as the beginning of the current chunk, where the marginal \(q(\vx_i)\) is explicitly represented and can be sampled directly. Moreover we can now ``minibatch'' over different chunks, evaluating our bound in a ``doubly stochastic'' manner \citep{salimbeni2017doubly}.


\subsection{Dependence on the inducing points \(\vu\) only - ``U-Factorisation''}
In order to avoid ``cutting'' long-ranging temporal dependences in our data by ``chunking'', we could instead use the inducing points to represent the dependence between \(\X\) and \(f\).
We can take \(q(\X\given f) = q(\X\given \vu)\) and, constraining \(q(\X\given \vu)\) to be Markovian (which is not exact in general, but is required for efficiency), we write:
\vspace{-0.2cm}
\begin{equation}
q(\vx_{t+1}\given \vu, \vx_t) = \NormDist{\xtp\given \at K_{\xt,\Z}K_{\Z,\Z}^{-1}\vu + \bt, \st + \at\cfut\at^\top}
\end{equation}
The intuition behind this choice of posterior (corresponding to the third one in the table) is to represent the GP with samples from the inducing point variational distribution (each sample effectively being a different transition function), and to generate trajectories \(\X\) for each of the samples. Because these functions are represented parametrically (with finite ``resolution'' corresponding to the number of inducing points), and our posterior is Markovian, our sampling complexity does not grow as we traverse the latent chain \(\X\). If we were to ``integrate out'' \(\vu\) (as in the second posterior in the table), however, the dependence between \(\X\) and \(f\) would be severed (recall \(\vu\) is ``part'' of \(f\)).


\section{Experiments}
\begin{figure}[!h]
\centering
\includegraphics[width=1.\linewidth]{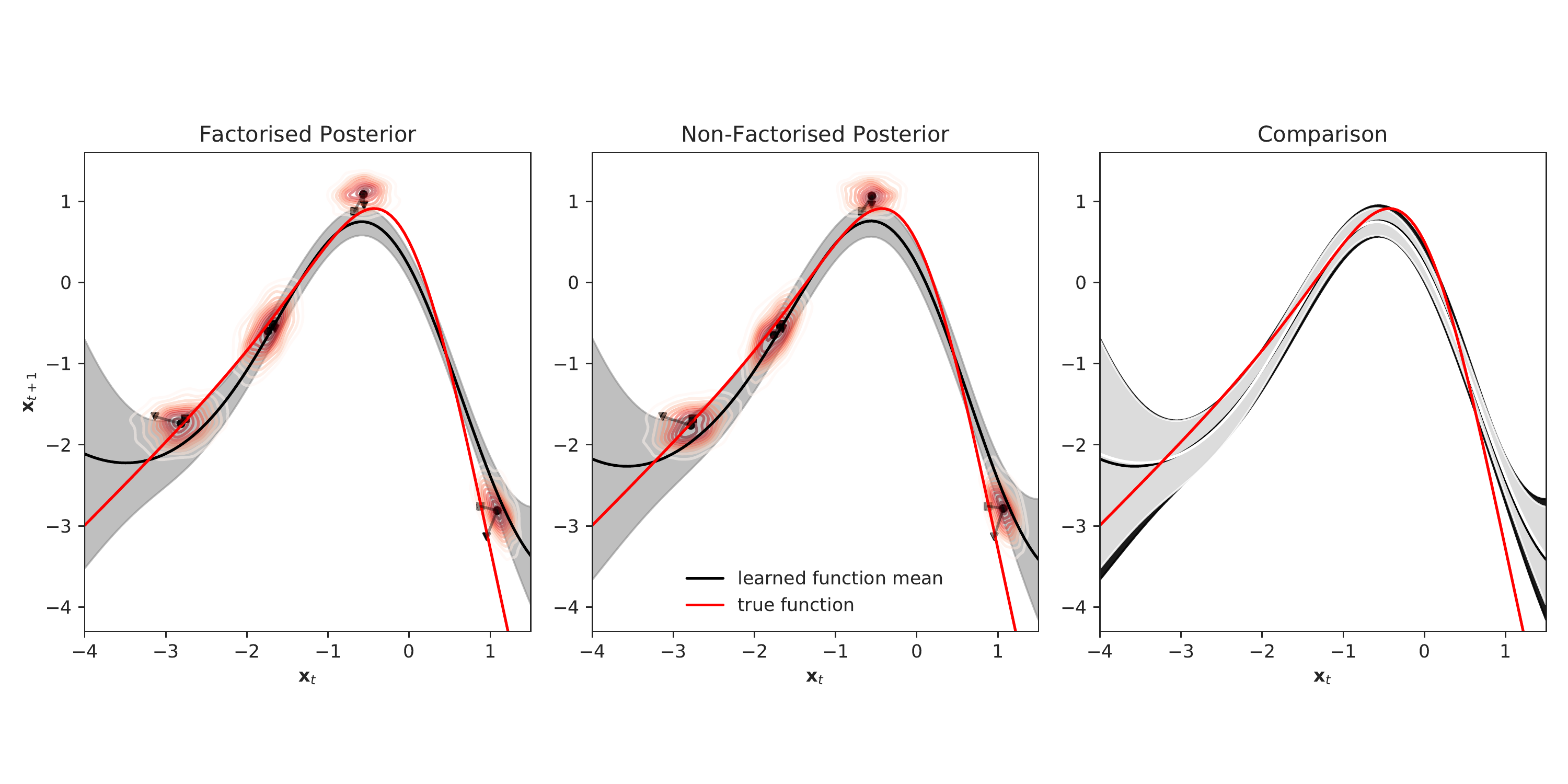}
\vspace{-1.5cm}
\caption{\small{Uncertainty shrinking due to factorisation. \textbf{Left}: fit for the factorised \(q(\X)q(f)\); \textbf{Middle}: fit for the U-Factorised \(q(\X\given \vu)q(f)\); \textbf{Right}: superimposed posteriors (grey is factorised, black is U-Factorised). \textbf{All subplots}: the shaded regions indicate a \(2\sigma\) confidence interval and the contour plots (left and middle) show the pairwise posteriors \(q(\xt, \xtp)\) over some test latent states. The \((x,y)\) coordinates of the squares, triangles and circles correspond to, respectively: the true latent states \((\xt,\xtp)\), the observed states \((\yt,\vy_{t+1})\) and the means of the marginal posteriors \((\Exp{q(\xt)}{\xt},\Exp{q(\xtp)}{\xtp})\).}}
\label{fig:kink}
\end{figure}

In order to test the effect of the factorisation on the approximate GP posterior (i.e. the learned dynamics), we perform inference on data\footnote{A sequence of 50 steps was generated with Gaussian emission and process noise (standard deviations of \(\sqrt{0.1}\) and \(\sqrt{0.01}\) respectively). The emission model was fixed to the generative one to allow comparisons.} generated by the ``kink'' transition function (see Figure \ref{fig:kink}'s ``true function''). The models whose fit is shown are ``Factorised - non-linear'' and ``U-Factorised - non-linear'', both using an RBF kernel. Factorisation leads to an overconfident, mis-calibrated posterior and this is the same for both factorised models (they gave virtually the same fit). Using 100 inducing points, the U-Factorised and Non-Factorised posteriors were also indistinguishable, the transition function being precisely ``pinned-down'' by the inducing points.

\pagebreak

\bibliography{references}

\appendix

\section{Comparison to PR-SSM}
\citet{doerr2018probabilistic} were, to the best of our knowledge, the first to consider a non-factorised variational posterior for the GPSSM\footnote{They call their approach PR-SSM.}. Their work, however, has two significant shortcomings. Firstly, the \(q(\xtp|f, \xt)\) terms are set to be the same as the corresponding prior terms (i.e. the prior transitions). This posterior fails to exploit information contained in the observations other than by adapting \(q(\vu)\) (it performs no filtering or smoothing on the latent states), and can only be an adequate approximation when the process noise is low and the observed sequence is short (as even low noise levels can compound, and potentially be amplified, in a long sequence). Of course, it would also be an appropriate choice (if somewhat difficult to optimise) when the process noise is zero, but then the latent variables become deterministic, given the transition function, and it is unclear whether modelling them explicitly through a probabilistic state-space model should be beneficial (an auto-regressive model with no latent variables might be sufficient). \\
Secondly, \citet{doerr2018probabilistic} employ a sampling scheme which gives incorrect marginal samples of \(q(\xt)\), even assuming, as they did, that:
\begin{equation}
    p(f(\vx_1), \dots,f(\vx_{T-1})|\vu) = \prod_{t=1}^{T-1} p(\ft|\bu) \label{fitc}
\end{equation}
The mistake is predicated on believing that the factorisation in equation \ref{fitc} produces a Markovian \(q(\X)\). In general, this is not the case. A mismatch is thus introduced between the form of the variational lower bound and the samples being used to evaluate it, resulting in a spurious objective.

\end{document}